\documentclass[journal]{IEEEtran}
\usepackage{amsmath}
\usepackage{amssymb}
\usepackage{color}
\usepackage{mathrsfs}

\usepackage{cite}

\usepackage[pdftex]{graphicx}

\ifCLASSINFOpdf
\else
\fi
\usepackage{algorithmic}

\begin{document}
%
\title{Show, Tell and Summarize: Dense Video Captioning\\ Using Visual Cue Aided Sentence Summarization}
%
%
%

\author{Zhiwang~Zhang,
        Dong~Xu,~\IEEEmembership{Fellow,~IEEE,}
        Wanli~Ouyang,~\IEEEmembership{Senior Member,~IEEE,}
        Chuanqi~Tan, 

\thanks{This research is supported by the Australian Research Council Future Fellowship under Grant FT180100116.(Corresponding author: Dong Xu.)}        
\thanks{Zhiwang Zhang, Dong Xu and Wanli Ouyang are with the School of Electrical and Information Engineering, University of Sydney, Sydney, NSW, 2008 Australia e-mail: (zhiwang.zhang@sydney.edu.au, dong.xu@sydney.edu.au, wanli.ouyang@sydney.edu.au).}
\thanks{Chuanqi Tan is with the Department of Computer Science and Technology, Tsinghua University, Beijing, China. e-mail: (chuanqi.tan@gmail.com)}
}


%
%

\markboth{IEEE TRANSACTIONS ON xxxx, ~Vol.XX, No.XX, XXXX,XXXX}%
{Shell \MakeLowercase{\textit{et al.}}: Bare Demo of IEEEtran.cls for Journals}
%



\maketitle

\begin{abstract}
In this work, we propose a division-and-summarization (DaS) framework for dense video captioning. After partitioning each untrimmed long video as multiple event proposals, where each event proposal consists of a set of short video segments, we extract visual feature (e.g., C3D feature) from each segment and use the existing image/video captioning approach to generate one sentence description for this segment. Considering that the generated sentences contain rich semantic descriptions about the whole event proposal, we formulate the dense video captioning task as a visual cue aided sentence summarization problem and propose a new two stage Long Short Term Memory (LSTM) approach equipped with a new hierarchical attention mechanism to summarize all generated sentences as one descriptive sentence with the aid of visual features. Specifically, the first-stage LSTM network takes all semantic words from the generated sentences and the visual features from all segments within one event proposal as the input, and acts as the encoder to effectively summarize both semantic and visual information related to this event proposal. The second-stage LSTM network takes the output from the first-stage LSTM network and the visual features from all video segments within one event proposal as the input, and acts as the decoder to generate one descriptive sentence for this event proposal. Our comprehensive experiments on the ActivityNet Captions dataset demonstrate the effectiveness of our newly proposed DaS framework for dense video captioning.
\end{abstract}


%
\IEEEpeerreviewmaketitle

\section{Introduction}
%
%
%
%

\IEEEPARstart{D}{ense} video captioning, which aims to generate a set of descriptive sentences for all events in an untrimmed long video, has recently attracted interesting attention in the computer vision community. Dense video captioning is a more challenging task than image and video captioning (See Section 2 for a brief survey of the recent progress in image and video captioning) as all events occurred in untrimmed long videos are expected to be located and described with natural language. As a result, dense video captioning technologies can benefit a broad range of applications such as video object detection, video retrieval and video summarization. 

Most existing dense video captioning approaches consist of two steps, in which temporal event proposals are first generated and then all detected events are described with natural language. In \cite{krishna2017dense},  Krishna et al. proposed to exploit contextual information from past and future events to generate better textual descriptions for all detected events, and also introduced the large benchmark dataset ActivityNet Captions for evaluating different dense video captioning methods. In \cite{wang2018bidirectional}, Wang et al. developed a bidirectional proposal method to generate more accurate event proposals by exploiting past and future contexts, and also proposed a new visual representation and a new context gating mechanism to better describe all events with natural language. To simultaneously detect and describe events with natural language, Li et al. \cite{li2018jointly} integrated a newly proposed descriptiveness component into a single shot detection framework, while Zhou et al. \cite{zhou2018end} also proposed a new transformer-based network  consisting of one encoder, one proposal decoder and one captioning decoder.  While these methods achieved promising results for the dense video captioning task, these works cannot work well when the scenes and objects in one event proposal are rapidly changing. 

In Section III, we first use a simple Long Short Term Memory (LSTM) based approach \cite{wang2018bidirectional} to generate event proposals for each untrimmed long video. Considering that each event proposal consists of a set of video segments, we extract visual feature (e.g., C3D feature) from each segment and then generate one sentence description for describing each segment by using any existing image/video captioning approach (e.g., \cite{xu2015show}). It is worth mentioning that the generated sentences provide rich semantic descriptions for the whole event proposal even when the scenes and objects in this event proposal are rapidly changing. We thus formulate the dense video captioning task as a visual cue aided sentence summarization problem, which summarizes all generated sentences as one descriptive sentence with the aid of visual features. In order to better capture temporal evolution of video clips within each event proposal, we propose a new two-stage LSTM based framework for dense video captioning, in which the two LSTM networks act as the encoder and decoder, respectively.  Using a set of semantic words from all generated sentences and the visual features from all video clips within one event proposal as the input, the first-stage LSTM network effectively summarizes both semantic and visual information related to one event proposal as intermediate hidden representation. Based on the output from the first-stage LSTM network and the visual features from all video clips within one event proposal, the second-stage LSTM network generates one descriptive sentence for this event proposal.

Considering that some video segments are more important than others for dense video captioning, we further propose a new hierarchical attention mechanism in the temporal domain.  In our framework, the LSTM module is related to one semantic word while a set of LSTM modules are related to the whole sentence from one segment. Therefore, two attention mechanisms learn the optimal weights for the visual and textual features from different levels.\color{black}
The newly proposed hierarchical attention mechanism can be readily incorporated into the two-stage LSTM network to further improve dense video captioning results. 

In Section IV, we conduct comprehensive experiments using the benchmark dataset ActivityNet Captions and the results clearly demonstrate our division-and-summarization (DaS) framework, which consists of a two-stage LSTM network equipped with the newly proposed hierarchical attention mechanism, outperforms the existing dense video captioning methods.

The major \textbf{contributions} of this work include: 1) after formulating the dense video captioning task as a new visual cue aided sentence summarization problem,  we propose a new DaS framework consisting of a two-stage LSTM network and a new hierarchical attention mechanism; 2) our DaS framework achieves promising results on the ActivityNet Captions dataset for dense video captioning.

\begin{figure}[t]
    \centering
    \includegraphics[width=\linewidth]{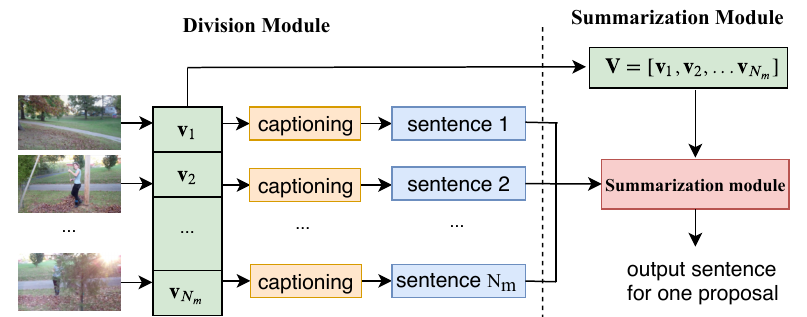}
    \vspace{-4mm}
    \caption{Overview of our division and summarization framework.}
    \label{overview}
\end{figure}

\section{Related Work}

Dense video captioning approaches are related to temporal action proposal methods \cite{duchenne2009automatic,caba2016fast,escorcia2016daps,buch2017sst,wang2018bidirectional} and image/video captioning methods \cite{shen2017weakly,pan2016jointly,chen2018less,wang2018video,wang2018reconstruction,wu2018interpretable,chen2018regularizing,su2019improving}, which will be discussed below. 

\subsection{Temporal action proposals}

The work in \cite{duchenne2009automatic} used the sliding window based approach to generate dense action proposals. The work in \cite{caba2016fast} proposed a faster method for detecting human actions in untrimmed videos. The Deep Action Proposals (DAPs)  method \cite{escorcia2016daps} incorporated the LSTM approach into the sliding window framework to encode the visual content in each sliding window and generate multi-scale action proposals. Building on the DAPs method, the single-stream temporal (SST) action proposals method \cite{buch2017sst} used non-overlapping sliding windows and encoded the visual content in each sliding window with the Gated Recurrent Unit (GRU) method. The bidirectional SST method \cite{wang2018bidirectional} further extended the SST method to encode both past and future contexts when localizing the current event.

\subsection{Image and video captioning}

The recent image/video captioning works used deep learning methods, in which the Convolutional Neural Networks (CNNs) are used as the encoder to encode images or video frames, while the LSTM networks are used as the decoder to generate descriptive languages \cite{li2018jointly}. Different frame encoding strategies including mean-pooling \cite{venugopalan2014translating,gan2017semantic}, attention mechanism\cite{yao2015describing,pan2017video,gan2017semantic} or recurrent networks \cite{sutskever2014sequence,donahue2015long,pan2016hierarchical} are employed in different methods. In contrast to these image/video captioning methods which can only handle images or short videos with only one major event, our work will generate a set of descriptive sentences for all events in each video \cite{krishna2017dense}. 

Different summarization strategies using frame-level captioning results were proposed for video captioning in \cite{li2015summarization,liu2016boosting}. But our work is different with the work \cite{li2015summarization,liu2016boosting} in the following two major aspects. 1) our work generates sentences for each video segment instead of using the image captioning method to generate sentences for each frame as in the existing approaches \cite{li2015summarization,liu2016boosting},  2) The work in  \cite{liu2016boosting} directly feeds the input word into LSTM without using visual features and any attention mechanism, while the work \cite{li2015summarization} uses LexRank \cite{erkan2004lexrank} to select the representative sentence with the highest score from the frame-level captioning results. In contrast to \cite{liu2016boosting,li2015summarization}, our summarization module considers the importance of each input word by using a newly proposed hierarchical attention mechanism, which is specially designed for our task. We also use visual features in our visual cue aided sentence summarization module.

Different attention strategies (e.g., spatial attention \cite{xu2016ask}, temporal attention \cite{song2017end}, semantic attention \cite{you2016image} or combination of multiple attention strategies \cite{yang2016stacked}), were proposed for different computer vision tasks including image classification \cite{wang2017residual}, object localization \cite{ba2014multiple} and image/video captioning \cite{xu2015show}. 

\color{black} In contrast to these works \cite{xu2016ask,song2017end,you2016image,yang2016stacked}, we propose a new hierarchical attention mechanism for dense video captioning which considers the group structure of the words from all output sentences after the division module. \color{black}

\section{Methodology}
An overview of our DaS framework for the dense video captioning task is shown in Figure \ref{overview}. Our work treats each proposal independently and generates one sentence description for each proposal. Specifically, our framework contains \textbf{three components}, the proposal generation module,  the division module and the summarization module. 
Correspondingly, our pipeline consists of three steps. 

1) The proposal generation module takes the visual features as the input and generates the candidate proposals that might contain actions. We utilize the Bi-AFCG method in \cite{wang2018bidirectional} to generate the event proposals in our implementation. 

2) Given the generated event proposals, the division module splits each video proposal into multiple segments. The video proposal is uniformly sampled into a predefined number of video segments and then we extract visual feature (e.g. C3D feature) for each segment.
The textual description for each segment is obtained by using the image/video captioning method \cite{vinyals2015show}. Details of this module are given in Section \ref{dm}. 

3) The summarization module summarizes the textual descriptions and visual features from all segments within each proposal to generate the final sentence description. This module is introduced in Section \ref{sm}.

\begin{figure*}[t]
    \centering
    \includegraphics[width=\textwidth]{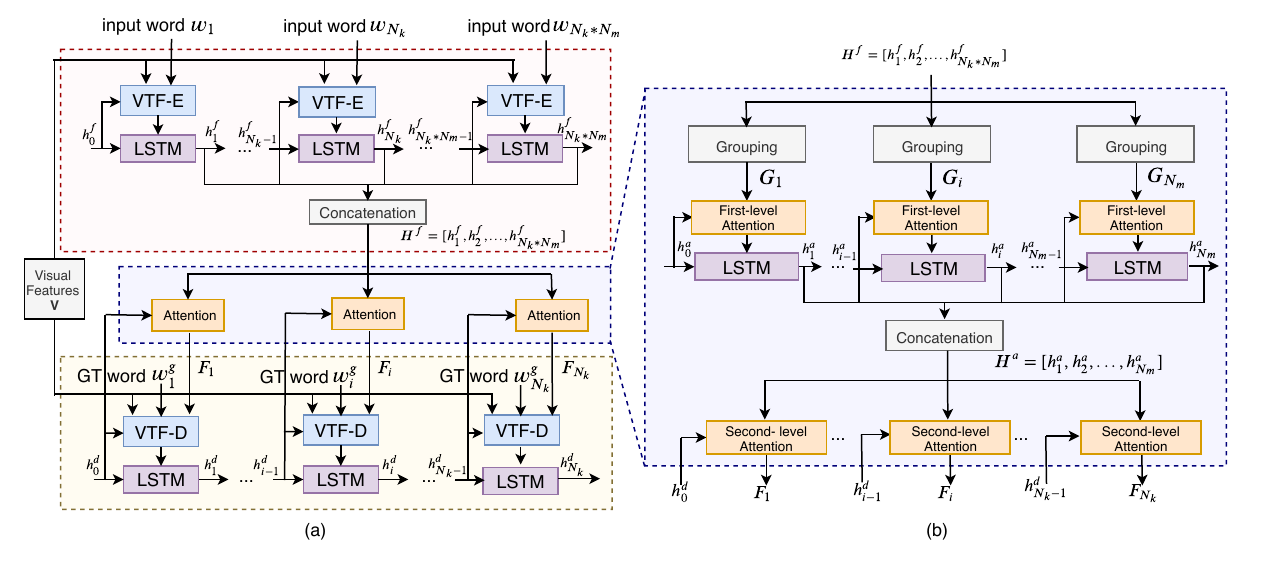}
    \vspace{-8mm}
    \caption{Overview of our summarization module (best viewed in color). All the operations are shown in the blocks with different colors. (a): our summarization module with simple-attention mechanism. The block with red dash line represents the encoder-fusion sub-net, the block with blue dash line represents the encoder-attention sub-net and the block with yellow dash line represents the decoder sub-net. ``GT word'' means ``ground-truth word''. (b): Encoder-attention sub-net with hierarchical-attention mechanism, which can be used to replace the encoder-attention sub-net in (a). } 
    \label{pipeline}
\end{figure*}

\begin{figure}[t]
    \centering
    \includegraphics[width = \linewidth]{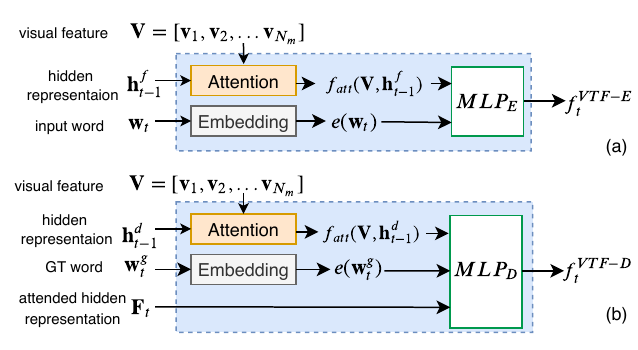}
    \vspace{-8mm}
    \caption{Illustration of the visual and textual feature fusion modules (a): at the encoder side (VTF-E) (b):at the decoder side (VTF-D) }
    \label{vtf}
\end{figure}

\subsection{Division module}\label{dm}
There are \textbf{two steps} in the division module (see Figure \ref{overview}): 1) sampling  a predefined number of segments for each event proposal and 2) video captioning.

In the sampling step, each event proposal is split into a set of segments with each segment consisting of 16 frames. 
We then uniformly sample $N_m$ video segments and extract visual feature (e.g. C3D feature) from these $N_m$ segments.  By default, $N_m$ is set to 20 in this work.  
In our implementation,  principal component analysis (PCA) is also utilized to reduce the dimension of each C3D feature from 4096 to 500. In this case, the visual feature $\textbf{v}_i$ for the $i$th segment is a 500 dimension feature vector. 

 In the video captioning step, we apply the existing captioning approach in  \cite{vinyals2015show} to generate the textual description by using each visual feature $\textbf{v}_i$ as the input. Each segment is described by a sentence for the corresponding event proposal. For fair comparison, only ActivityNet Captions Dataset is used in the captioning step and extra datasets such as MSCOCO and MSVD datasets are not used. 

In summary, for each generated proposal, our division module produces the following $N_m$ visual features and sentences which are used as the input for the summarization module.

\begin{equation}
  \textrm{Output of the division module}:   \left\{
    \begin{aligned}
    \textbf{V} &= \{\textbf{v}_1, \textbf{v}_2, ..., \textbf{v}_{N_m} \} \\
    \textbf{S} &= \{\textbf{s}_1, \textbf{s}_2, ..., \textbf{s}_{N_m} \} \\
    \end{aligned},
    \right. \label{eq:splitter}
\end{equation}
where $\textbf{v}_t$ is the visual feature for the $t$th segment and $\textbf{s}_t$ is the generated sentence of this segment.  Each sentence $\textbf{s}_t$ contains $N_k$ words $\textbf{s}_t = \{\textbf{w}_{(t-1)*N_k+1}, ..., \textbf{w}_{t*N_k}\}, t = 1,2,...N_m$.

\subsection{Summarization module}\label{sm}

Our summarization module has \textbf{three sub-nets}: encoder-fusion sub-net, encoder-attention sub-net and decoder sub-net. The encoder-fusion sub-net fuses the textual features and visual features for all segments and produces the hidden representation for each input word.  The encoder-attention sub-net summarizes the hidden representation from the encoder-fusion sub-net and outputs the attended hidden representation. The decoder sub-net takes the attended hidden representation from the encoder-attention sub-net and visual feature from all segments to generate the sentences for the proposal.

Firstly, we introduce the $LSTMCell(\cdot)$ \cite{hochreiter1997long} which is used as the basic block in our model. The $LSTMCell(\cdot)$ is defined as:
\begin{equation}
\begin{pmatrix}\textbf{i}_t\\ \textbf{f}_t \\\textbf{o}_t\\\textbf{g}_t\end{pmatrix} =
  \begin{pmatrix}\sigma\\\sigma\\\sigma\\\tanh\end{pmatrix}
  \textbf{U}_{LSTM}\begin{pmatrix}\textbf{z}_{t}\\\textbf{h}_{t-1}\end{pmatrix}\\
\end{equation}

where $\textbf{i}_t,\textbf{f}_t,\textbf{o}_t$ are the input gate, the forget gate and the output gate at time step t respectively. $\sigma$ is the sigmoid function. 
$\textbf{h}_t $ is the hidden state of the LSTMCell, 
$\textbf{z}_t$ is the context vector at time step t. $\textbf{z}_t$ may be the weighted sum of visual features and/or textual features depending on the position of the LSTMCell. $\textbf{U}_{LSTM}$ is a transformation matrix to be learned. The memory cell $\textbf{c}_t$ and the hidden state $\textbf{h}_t$ are defined as:
\begin{equation}
\begin{split}
\textbf{c}_t &= \textbf{f}_t \odot \textbf{c}_{t-1} + \textbf{i}_t \odot \textbf{g}_t\\
\textbf{h}_t &= \textbf{o}_t \odot \tanh(\textbf{c}_t)
\end{split}
\end{equation}
where $\odot$ denotes the element-wise multiplication operator.

\emph{LSTM with attention} will be used by all the components in the summarization module, which is formulated as follows:
\begin{equation}
\textbf{h}_t = LSTMCell(f_{att}(\textbf{X},\textbf{h}_{t-1}), \textbf{h}_{t-1}),
\end{equation}
where $\textbf{X}=\{\textbf{x}_1, \ldots, \textbf{x}_n, \ldots, \textbf{x}_N\}$ denotes the set of input features with total length of N, $\textbf{h}_t$ and $\textbf{h}_{t-1}$ denote the hidden state at time step $t$ and $t-1$, respectively. $f_{att}(\textbf{X},\textbf{h}_{t-1})$ denotes the attention function \cite{yao2015describing} defined as follows:
\begin{equation} \label{eq:att}
\begin{split}
f_{att}(\textbf{X},\textbf{h}_{t-1}) &= \sum\limits_{i=1}^N a_{t, i} * \textbf{x}_i, \\
\textrm{where } a_{t, i} &= \dfrac{\exp(MLP(\textbf{x}_i, \textbf{h}_{t-1}))}{\sum\limits_{j=1}^N \exp(MLP(\textbf{x}_j, \textbf{h}_{t-1}))}, \\
\end{split}
\end{equation}
where $MLP(\cdot, \cdot)$ is a multi-layer perceptron function, $a_{t, i}$ is the learned attention weight for the $i$th input feature $\textbf{x}_i$. The attention weight is calculated from the input features $\textbf{X}$ and the hidden state $\textbf{h}_{t-1}$ for the LSTM cell.

\subsubsection{Encoder-fusion sub-net} 
The encoder-fusion sub-net takes the words $\textbf{S}$ and visual features $\textbf{V}$ in Eq. (\ref{eq:splitter}) from the whole proposal as the input and obtains the hidden representation by using LSTM with attention mechanism. 

Denote the number of words in $\textbf{S}$ by $|\textbf{S}|$, in which $|\textbf{S}|=N_m*N_k$. Denote the ${t}$th word from the set $\textbf{S}$ by $\textbf{w}_t$, where $t = 1 \ldots, |\textbf{S}|$. As shown in Figure \ref{vtf} (a), the textual and visual feature fusion module at the encoder side (VTF-E)  fuses both visual features and textual features together, which is defined below:
\begin{equation}
f^{VTF-E}_t = {MLP_E}(f_{att}(\textbf{V},\textbf{h}^f_{t-1}), e(\textbf{w}_{t}))
\label{eq:VTF-E}
\end{equation}
where $f_{att}(\textbf{V},\textbf{h}^f_{t-1})$ is the attended visual feature and $\textbf{h}^{f}_{{t}-1}$ is the hidden representation at time step ${t}-1$,  which is defined in Eq. (\ref{eq:f-lstm}). $e(\textbf{w}_{t})$ is the embedded feature vector for the ${t}$th word $\textbf{w}_{t}$ through the embedding function. ${MLP_E}(\cdot,\cdot)$ is a multi-layer perceptron function. According to the formulation in Eq. (\ref{eq:VTF-E}), the words from the division module are transformed into the embeded feature 
and then combined with the attended visual features 
before being fed into the LSTM cell.

The hidden representation at time step ${t}$ of the encoder-fusion sub-net, denoted by $\textbf{h}^{f}_{t}$, is obtained from the input words and visual features as follows:

\begin{equation}
\textbf{h}^{f}_{t} = LSTMCell( f^{VTF-E}_t,  \textbf{h}^{f}_{{t}-1}),
\label{eq:f-lstm}
\end{equation}
where $f^{VTF-E}_t$ is the output of the VTF-E module at time step $t$. The LSTM module takes all visual and textual features within each event proposal into consideration. In this way, the hidden representation at the current time step contains not only the fused information from the current word, but also the information from the previous words and the visual information from all segments.
The hidden representation of all times steps are concatenated as the output of the encoder-fusion sub-net: ${\textbf{H}^f} = \{\textbf{h}^f_1, \textbf{h}^f_2, ..., \textbf{h}^f_{|\textbf{S}|}\}$.

\subsubsection{Encoder-attention sub-net}

The hidden representation from the encoder-fusion sub-net are used as the input of the encoder-attention sub-net. There are \textbf{two types} of attention  mechanism can be used in our framework as encoder-attention sub-net: simple-attention mechanism and hierarchical-attention mechanism.

\textbf{Simple-attention mechanism}:
 The simple-attention mechanism is utilized to summarize the hidden representation from the encoder-fusion sub-net before being feeding into the decoder sub-net. $\textbf{F}_{{t}}$ is the attended hidden representation defined as follows:

\begin{equation}
\label{eq:r}
\textbf{F}_{{t}} = f_{att}(\textbf{H}^f, \textbf{h}^{d}_{{t}-1}),
\end{equation}
where $f_{att}(\cdot,\cdot)$ is the attention function defined in Eq. (\ref{eq:att}), $\textbf{h}^d_{t-1}$ is the hidden state of the decoder at time step $t-1$ defined in Eq. (\ref{eq:decoder}) and ${\textbf{H}^f} = \{\textbf{h}^f_1, \textbf{h}^f_2, ..., \textbf{h}^f_{|\textbf{S}|}\}$ is the concatenated hidden representation from the encoder-fuse sub-net.

\textbf{Hierarchical-attention mechanism}: 
Considering that the group structure of input words that belong to different sentences, we propose a hierarchical attention mechanism. 
The input features of the hierarchical-attention mechanism are the same as the input features of the simple-attention mechanism. 
\color{black} However, the LSTMs with a newly proposed two-level attention mechanism are used to summarize the hidden representation from the encoder-fusion sub-net before being fed into the decoder sub-net.\color{black} 

Specifically, the hidden representations ${\textbf{H}^f}$ from the encoder-fusion sub-net are first grouped into $N_m$ groups of the features, in which each group of features corresponds to one sentence from each segment. Denote the features for the ${t}$-th segment by $\textbf{G}_{t}$ where $\textbf{G}_{t}=\{\textbf{h}^f_{({t}-1)N_k+1}, ..., \textbf{h}^f_{{t} \times N_k}\}, t = 1,...,N_m$. 

The hidden representations from the encoder-attention sub-net, which represents the information for the ${t}$-th segment is denoted by $\textbf{h}^{a}_{t}$. It is obtained from the grouped features $\textbf{G}_{t}$ of the ${t}$th segment below: 
\begin{equation}
\textbf{h}^{a}_{t} = LSTMCell(f_{att}(\textbf{G}_{t}, \textbf{h}^{a}_{{t}-1}), \textbf{h}^{a}_{{t}-1} ),
\label{eq:hs}
\end{equation}
where $\textbf{G}_{{t}}$ is the grouped futures for the time step $t$ and $f_{att}(\cdot,\cdot)$ is the attention function defined in Eq. (\ref{eq:att}). These hidden representations, which represent the hidden features for the $N_m$ segments, are then concatenated as $\textbf{H}^{a} = \{\textbf{h}^{a}_{1},...\textbf{h}^{a}_{N_m}\}$.

$\textbf{F}_{{t}}$ is the attended hidden representation fed into the visual and textual feature fusion module at the decoder side (VTF-D) at time step $t$ which is defined as follows:
\begin{equation}
\textbf{F}_{{t}} = f_{att}(\textbf{H}^a, \textbf{h}^{d}_{{t}-1}),
\end{equation}
where $f_{att}(\cdot,\cdot)$ is the attention mechanism defined in Eq. (\ref{eq:att}), $\textbf{h}^d_{t-1}$ is the hidden state of decoder at time step $t-1$ defined in Eq. (\ref{eq:decoder}) and $\textbf{H}^a$ is the hidden representations.

\subsubsection{Decoder sub-net}  
After the attended hidden representation $\textbf{F}_t$ is obtained from the encoder-attention sub-net, the decoder sub-net uses them to generate one sentence description for each proposal.

In the decoder, we use LSTM to generate one sentence description. The textual and visual feature fusion module at the decoder side (VTF-D) is shown in Figure \ref{vtf} (b). In this phase, we use VTF-D to fuse the visual features $\textbf{V}$, the ground-truth word $\textbf{w}^g_t$ for the decoder at time step $t$ and
 the intermediate hidden representation ${\textbf{F}_{t}}$ from the encoder-attention sub-net at time step t as follows:
\begin{equation}
f^{VTF-D}_t = MLP_D(f_{att}(\textbf{V},\textbf{h}^d_{t-1}), e(\textbf{w}^g_{t}), \textbf{F}_{t})
\label{eq:VTF-D}
\end{equation}
where $f_{att}(\textbf{V},\textbf{h}^d_{{t}-1})$ 
is the attended visual feature and $\textbf{h}^{d}_{{t}-1}$ is the hidden representation of the decoder at time step ${t}-1$ which is defined in Eq. (\ref{eq:decoder}). $e(\textbf{w}^g_{t})$ is the embedded feature vector for the ${t}$th word $\textbf{w}^g_t$ through the embedding function. The input word $\textbf{w}^g_t$ represents the ground-truth of the previous word or the generated previous word in the training stage and testing stage, respectively. $MLP_D(\cdot,\cdot,\cdot)$ is a multi-layer perceptron function. 

The hidden representation of the decoder at time step $t$ is obtained as follows: 
\begin{equation}
\textbf{h}^d_t = LSTMCell (f^{VTF-D}_t,\textbf{h}^d_{t-1})
\label{eq:decoder}
\end{equation}

where $f^{VTF-D}_t$ is the output of the VTF-D module at time step $t$, $\textbf{h}^d_{{t}}$ is the hidden representation of the decoder, which is finally used to generate the ${t}$th word of the output sentence.

The  hidden representation in decoder $\textbf{h}^d_{t}$ are then used for estimating the ${t}$th word $\textbf{w}^o_t$ of the generated sentence as follows:
\begin{equation}
\begin{split}
\textbf{w}^o_t = softmax(\textbf{U}_o \textbf{h}^d_{t})
\end{split},
\end{equation}
where $\textbf{U}_o$ is the transformation matrix to be learned and $softmax(\cdot)$ is the softmax function.
$\textbf{w}^o_t = [{w}^o_{t,1},{w}^o_{t,2}, ... ,w^o_{t, N_v}]$, where $w^o_{t, m}$ represents the probability of the $m$th word in the dictionary and the number of words in the dictionary is denoted by $N_v$.

\subsection{Other Details}
The final loss function is defined as follow:
\begin{equation}\label{eq:loss}
    \mathcal{L} = \mathcal{L}_{cross-entropy} +\lambda_{d} \mathcal{L}_{d},
\end{equation}
where $\lambda_d$ is the hyper-parameter. $\mathcal{L}_{cross-entropy}$ is the commonly used cross entropy loss from \cite{vinyals2015show}, and $\mathcal{L}_{d}$ is the discriminative loss \cite{yang2016review} that guides the model to predict discriminative objectives (word occurrences) in the encoder-fusion sub-net. 

In order to improve the performance of dense video captioning, we follow the work \cite{rennie2017self} and apply reinforcement learning in our framework, in which we use Meteor as the reward rather than CIDEr-D. 

\section{Experiments}
\begin{table*}[ht]
\begin{minipage}{1\linewidth}
\centering
\caption{Performance of different methods on the ActivityNet captions validation dataset when using the ground-truth proposals.}
\begin{tabular}{|c|c|c|c|c|c|c|c|}
\hline
Model  & BLEU1 & BLEU2 & BLEU3 & BLEU4 & Rouge-L & Meteor & CIDEr-D \\
\hline
LSTM \cite{venugopalan2014translating}  & 18.40 & 8.76 & 3.99 & 1.53 & - & 8.66 & 24.07 \\
\hline
S2VT \cite{venugopalan2015sequence}  & 18.25 & 8.68 & 4.02 & 1.57 & - & 8.74 & 24.05 \\
\hline
TA \cite{yao2015describing}  & 18.19 & 8.62 & 3.98 & 1.56 & - & 8.75 & 24.14 \\
\hline
H-RNN \cite{yu2016video}  & 18.41 & 8.80 & 4.08 & 1.59 & - & 8.81 & 24.17 \\
\hline
DCE \cite{krishna2017dense}  & 18.13 & 8.43 & 4.09 & 1.60 & - & 8.88 & 25.12 \\
\hline
DVC \cite{li2018jointly} & 19.57 & 9.90 & \textbf{4.55} & 1.62 & - & 10.33 & 25.24 \\
\hline
DaS (Ours)  & \textbf{22.76} & \textbf{10.12} & 4.26 & \textbf{1.64} & \textbf{22.85} & \textbf{10.71} & \textbf{31.41} \\
\hline
\end{tabular}
\label{table:gt}
\end{minipage}
\end{table*}

\begin{table*}[ht]
\begin{minipage}{1\linewidth}
\centering
\caption{Performance of different methods on the ActivityNet Captions validation dataset when using automatically generated  proposals.}
\begin{tabular}{|c|c|c|c|c|c|c|c|}
\hline

 Model  & BLEU1 & BLEU2 & BLEU3 & BLEU4 & Rouge-L & Meteor & CIDEr-D \\
\hline
MT \cite{zhou2018end}   & -- & -- & \textbf{4.76} & 2.23 & - & 9.56 & -- \\
\hline
Bi-AFCG \cite{wang2018bidirectional} & 18.99 &8.84 & 4.41 & \textbf{2.30} & 19.10 & 9.60 & 12.68 \\ \hline

 DaS (Ours)  & \textbf{21.05} & \textbf{9.84} & 4.69 & 2.09 & \textbf{21.21} & \textbf{10.33} & \textbf{12.93} \\
\hline
\end{tabular}
\label{table:learned}
\end{minipage}
\end{table*}

\begin{table}[ht]
\begin{minipage}{1\linewidth}
\centering
\caption{Performance of our method and the baseline method Bi-AFCG \cite{wang2018bidirectional} for two groups of videos with different number of shot boundaries on the ActivityNet Captions dataset.}
\begin{tabular}{|c|c|c|c|c|}
\hline
Method &  $\#$ shot boundaries &  $\#$ videos & Meteor & Cider-D \\
\hline
 Bi-AFCG \cite{wang2018bidirectional}  & $<$ 10 & 2513 & 9.73 & 13.78 \\
\cline{2-5} 
& $\geqslant$10 & 2404 & 9.46 & 11.52 \\
\hline
DaS(Ours)  &$<$10 & 2513 & 10.39 & 14.01  \\
\cline{2-5}
  & $\geqslant$ 10 & 2404 & 10.10  & 13.26 \\ 
\hline
\end{tabular}
\label{table:shot_perform}
\end{minipage}
\end{table}

\begin{table}[ht]
\begin{minipage}{1\linewidth}
\centering
\caption{Performance of different methods on the ActivityNet Captions 2018 online testing server.}
\begin{tabular}{|c|c|}
\hline
Model  & Meteor\\
\hline
Bi-AFCG \cite{wang2018bidirectional} & 4.99  \\
\hline
DaS (Ours) & 6.40  \\
\hline
\end{tabular}
\label{table:test}
\end{minipage}
\end{table}

\subsection{Dataset}
The ActivityNet Captions dataset \cite{krishna2017dense} is a dense video captioning dataset with temporal event information.
The videos of this dataset are based on ActivityNet V1.3 \cite{caba2015activitynet}, which contains around 20k untrimmed videos from Youtube. 
The train/validation/test sets contain around 10k/5k/5k videos respectively. Each video is annotated with several human-written sentences to describe the events. Specifically, there are 3.65 sentences for each video on average and over 100,000 sentences for the total dataset.

\textbf{Data Preprocessing}
First, we discard all non-alphabetic characters. Then, we convert all the characters to lowercase. Finally, in order to control the size of the vocabulary in the dictionary, we replace all the sparse words appearing less than three times in the training set with the tag $<UNK>$.  Therefore, the dictionary eventually has 6,994 words.

\subsection{Experimental Setup}
For video representation, we follow the work in \cite{krishna2017dense,li2018jointly,wang2018bidirectional} and use the 500-dimension C3D feature \cite{krishna2017dense}.
This 500-dimension C3D feature is not fine-tuned on ActivityNet and the dimension is reduced from 4096 to 500 by using PCA.

\color{black} We set the number of sentences $N_m$ to 20 and the max length of words for each sentence $N_k$ to 25. Only the first 25 words will be remained for the sentences containing more than 25 words.  Zero padding is applied for pre-processing when the number of words in the generated sentences is less than 25. \color{black}
For sentence summarization, we make sure that the length of the total input words is fixed as 500.
The hyper-parameter $\lambda_{d}$ for the discriminative loss is empirically set to 0.1.

We implement our model based on PyTorch and use Adam \cite{kingma2014adam} as the optimizer. The initial learning rate for the cross-entropy loss  is set to 0.0003.
The learning rate will decrease by 1.25 times every 3 epochs.
The mini-batch size is set as 32. The hidden state size and the dropout rate for all LSTM cells is 512 and 0.8, respectively. \color{black} In our work, the multi-layer perceptron functions are implemented by using two fully-connected layers. \color{black}

\begin{table}[ht]
\begin{minipage}{1\linewidth}
\centering
\caption{Performance of different variants of our method on the ActivityNet Captions dataset.}
\begin{tabular}{|c|c|c|}
\hline
Model  & Meteor & CIDEr-D \\
\hline
TA (our implementation) \cite{yao2015describing} & 9.14 & 29.60 \\
\hline
DM-ave   & 8.46 & 24.52 \\
\hline
DM-best  & 9.07 & 28.91 \\
\hline
DaS(SA)  & 9.30 & 29.58 \\
\hline
DaS(HA)  & 9.55 & 30.47 \\
\hline
DaS (Ours)  & 10.71 & 31.41 \\
\hline
\end{tabular}
\label{table:ablation1}
\end{minipage}
\end{table}

\begin{table}[ht]
\begin{minipage}{1\linewidth}
\centering
\caption{Performance of different variants of our method on  the ActivityNet Captions dataset with/without visual feature.}
\begin{tabular}{|c|c|c|}
\hline
Model   & Meteor & CIDEr-D \\
\hline
DaS(SA) w/o VF-E   & 9.17 & 27.72 \\
\hline
DaS(SA) w/o VF-D   & 9.20 & 27.66 \\
\hline
DaS(SA) w/o VF-ED  & 9.13 & 27.85 \\
\hline
DaS (SA)    & 9.30 & 29.58 \\
\hline
\end{tabular}
\label{table:visual ablation}
\end{minipage}
\end{table}

\begin{table}[ht]
\begin{minipage}{1\linewidth}
\centering
\caption{Performance of our method when using different values of $\lambda_{d}$ on the ActivityNet Captions dataset.}
\begin{tabular}{|c|c|c|}
\hline
$\lambda_{d}$ value   & Meteor & CIDEr-D \\
\hline
0  & 9.45 & 29.57 \\
\hline
0.01  & 9.51 & 30.11 \\
\hline
0.1  & 9.55 & 30.47 \\
\hline
1    & 9.51 & 30.31 \\
\hline
\end{tabular}
\label{table:lambda}
\end{minipage}
\end{table}

\begin{table}[ht]
\begin{minipage}{1\linewidth}
\centering
\caption{Performance of our method when using different number of proposals (i.e.,$N_m$) on the ActivityNet Captions dataset.}
\begin{tabular}{|c|c|c|}
\hline
$N_m$   & Meteor & CIDEr-D \\
\hline
10  & 9.51 & 30.11 \\
\hline
20  & 9.55 & 30.47 \\
\hline
40  & 9.52 & 30.33 \\
\hline
\end{tabular}
\label{table:proposal_number}
\end{minipage}
\end{table}

\subsection{Baselines and Evaluation Metrics}

\textbf{Baselines.}
In order to evaluate the performance of the proposed approach, 
we follow the evaluation method in \cite{krishna2017dense} and compare our method with the following baseline methods.

(1) Translating Video to Natual Language (LSTM) \cite{venugopalan2014translating}. It adopted the basic CNN-RNN model for video captioning.
(2) Sequence to Sequence, Video to Text (S2VT) \cite{venugopalan2015sequence}. It used both RGB and optical flow as the visual inputs for video captioning.
(3) Temporal Attention (TA) \cite{yao2015describing} introduced the attention mechanism into video captioning.
(4) Hierarchical Recurrent Neural Networks (H-RNN) \cite{yu2016video} proposed a hierarchical RNN architecture to generate one paragraph with several sentences.
(5) Dense-Captioning Events (DCE) \cite{krishna2017dense} used DAPs \cite{escorcia2016daps} to localize temporal event proposals and S2VT \cite{venugopalan2015sequence} as the basic captioning module. Attention mechanism is used to explore the neighbouring event information.
(6) Dense Video Captioning (DVC)\cite{li2018jointly} introduced the descriptiveness-driven temporal attention mechanism, attributes and reinforcement learning into the captioning module.
(7) Bidirectional Attentive Fusion with Context Gating (Bi-AFCG)\cite{wang2018bidirectional} used a context gating approach with attention mechanism for the sentence generation module. Joint ranking is also employed to select sentences.
(8) Mask Transformer (MT) \cite{zhou2018end} proposed an end-to-end transformer model with a attention mechanism for dense video captioning.

The first four methods were proposed for the general video captioning task. The work in \cite{li2018jointly} also extended their methods for dense video captions.
All the baseline methods use C3D features except the work in \cite{zhou2018end}, which uses Resnet200 and optical flow as the visual features.

\textbf{Evaluation Metrics.} 
To evaluate different dense video captioning methods, we follow the work in  \cite{krishna2017dense,wang2018bidirectional,li2018jointly,zhou2018end} and use the mean of BLEU{1-4} \cite{papineni2002bleu}, Rouge-L \cite{lin2004rouge}, Meteor \cite{denkowski2014meteor} and CIDEr-D \cite{vedantam2015cider} across multiple intersection over union (IoU) thresholds ( 0.3, 0.5, 0.7 and 0.9 ) to measure the performance on the validation set.
For the testing set, only mean of Meteor is provided by the online testing server of ActivityNet Captions Challenge 2018.

\begin{figure}[t]
    \centering
    \includegraphics[width=\linewidth]{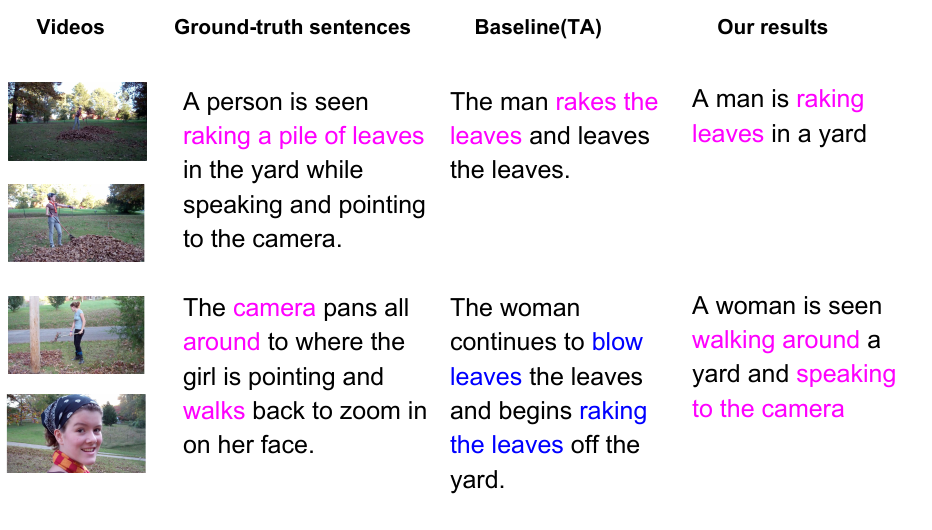}
    \vspace{-10mm}
\caption{Generated sentences by our method \textbf{DaS(HA)} and the baseline method \textbf{TA} \cite{yao2015describing} for one video from the Activity Captions validation set.  
     Correctly captured actions are highlighted by purple color, while incorrectly captured actions are highlighted by blue color.}
    \label{qa}
\end{figure}

\subsection{Performance Comparison}
We compare our proposed DaS method with the baseline methods on the ActivityNet Captions dataset. The results on the validation set based on the ground-truth proposals and automatically generated event proposals are shown in Table \ref{table:gt} and Table \ref{table:learned} respectively.
The result returned by the online testing server is reported in the Table \ref{table:test}.

\color{black}
In addition, in order to investigate the influence of the rapidly changing scenes and objects in ActivityNet, we compare our proposed model with the baseline method Bi-AFCG using the automatically generated event proposals in Table \ref{table:shot_perform}.
\color{black}

In Table \ref{table:gt}, our proposed method achieves the best performance in term of most evaluation metrics when using the C3D feature as the visual feature and the ground truth proposals. 
In particular, the results of our method are 10.71\% (Meteor score) and 31.41\% (CIDEr-D score), which outperform the previous state-of-the-art algorithm \cite{li2018jointly} by 0.38 \% (Meteor score) and 6.17\% (CIDEr-D score), respectively.
It is worth mentioning that DVC \cite{li2018jointly} utilized activity category labels as attributes while we do not use them.

As shown in Table \ref{table:learned}, our method also achieves the state-of-the-art performance in terms of Meteor score and CIDEr-D score, when using the C3D features as the visual features and the automatically generated event proposals.
For fair comparison, we use the same proposals as in the Bi-AFCG method in \cite{wang2018bidirectional}. 
Though our BLEU4 score is slightly lower than the Bi-AFCG method, our proposed model perform better than the Bi-AFCG method in term of  other indicators. Specifically, the Meteor score of our method is 10.33, while the Meteor score from Bi-AFCG method is only 9.60. 
\color{black} For dense video captioning, the Meteor score is a more important indicator, so we can claim that our overall performance is better than the Bi-AFCG method in \cite{wang2018bidirectional}.  \color{black}
Although the MT method in \cite{zhou2018end} uses both optical flow and Resnet200 features as the visual features, our method still achieves better Meteor score by using the C3D feature only,
which demonstrates the effectiveness of our framework.

We also submitted our results to the ActivityNet Captions Challenge 2018 , in which we only using the C3D feature and without ensembling multiple models.
Our proposed approach achieves better result than the Bi-AFCG method \cite{wang2018bidirectional} in term of the Meteor score. 

In order to compare our method with the baseline method Bi-AFCG for the videos with relatively fast or slow scene changes, we use the open-source library FFmpeg to detect the shot boundaries and group the videos in the validation set as two groups based on whether there are no less than 10 shot boundaries. 
When the number of shot boundaries is large (say no less than 10), there are more rapidly changing scenes and objects in these videos.
In Table \ref{table:shot_perform}, we compare our method with the baseline method Bi-AFCG for the videos in each group. For the videos with no less than 10 shot boundaries, our method achieves much better results than the baseline method. The results clearly demonstrate the effectiveness of our proposed method for the videos having rapidly changing scenes.

\begin{figure}[t]
    \centering
    \includegraphics[scale = .46]{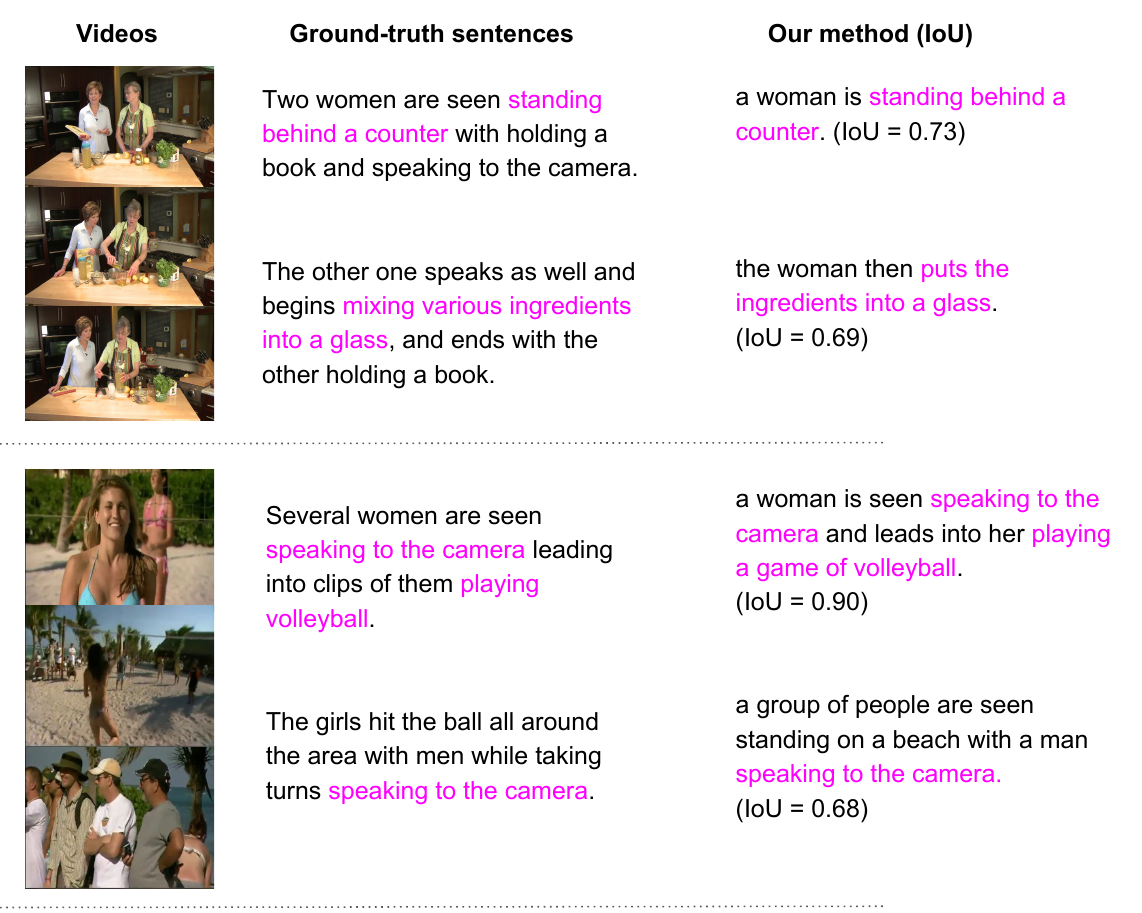}
    \vspace{-15pt}
    \caption{Illustration of generated sentences by using automatically generated proposals, in which the IoU value between each automatically generated proposal and the ground-truth proposal is also reported. Correctly captured actions are highlighted by purple color.} 
    \label{visual2}
\end{figure}

\subsection{Component Analysis}
In Table \ref{table:ablation1}, we conduct the following ablation experiments to verify the effectiveness of each newly proposed component in our network. Specifically, we report the results of the following methods.

(1) \textbf{TA} (our implementation): we re-implement the baseline temporal-attention method\cite{yao2015describing}. It consists of the decoder sub-net of our DaS network only, without generating multiple sentences. Therefore, the VTF-D module only takes the visual feature $\textbf{V}$ and the textual feature $\textbf{w}^{g}_t$ as the input in our VTF-D module. 
(2) \textbf{DM-ave}: We first use our division module to generates $N_m$ sentences for each event proposal, and then calculate a set of scores by comparing each generated sentence with the ground-truth sentence, and finally report the average score.
(3) \textbf{DM-best}: We first use our division module to generate $N_m$ sentences for each event proposal. We then calculate the confident score defined as $ \frac{1}{M} \sum\limits_{i=1}^M \log (p(\textbf{w}_i))$, where $p(\textbf{w}_i)$ indicates the probability of each predicted word in our approach and $M$ represents the number of words in the generated sentences. We then select the sentence with the highest confident score which is finally compared with the ground-truth sentence to calculate the score.
(4) \textbf{DaS(SA)}: Our DaS approach with the simple-attention mechanism. 
(5) \textbf{DaS(HA)}: Our DaS approach with the hierarchical-attention mechanism. 
(6) \textbf{DaS(Ours)}: Our DaS approach with the hierarchical-attention mechanism and reinforcement learning.

From the results in Table \ref{table:ablation1}, we have the following observations. First, our newly proposed approach DaS with simple-attention mechanism \textbf{DaS(SA)} outperforms the baseline model \textbf{TA}, 
which demonstrates that it is useful to first generate a set of sentences for all video segments and then summarize them into one sentence.
Second, our approach \textbf{DaS(SA)} also outperforms both \textbf{DM-ave} and \textbf{DM-best}, which indicates that it is beneficial to use our newly proposed sentence summarization method with the aid of visual features to generate more descriptive sentences for each event proposal. Third, our DaS approach with hierarchical-attention mechanism \textbf{DaS(HA)} is better than our DaS with simple attention mechanism \textbf{DaS(SA)}, which demonstrates that it is helpful to incorporate our newly proposed hierarchical attention mechanism in our DaS framework for dense video captioning. Finally, the final result from \textbf{DaS(Ours)} is the best, which indicates that the result from dense video captioning can be further improved by using reinforcement learning \cite{rennie2017self}.

In addition, in order to verify the effectiveness after using the visual features as the input in our DaS framework and the attention mechanism in both VTF-E module and VTF-D modules, we take our DaS framework with simple attention mechanism \textbf{DaS(SA)} as an example to perform the following ablation experiments in Table \ref{table:visual ablation}.

(1) \textbf{DaS(SA) w/o VF-E}: Our DaS approach without using visual features as the input in the encoder only. Specifically we do not use the visual feature $\textbf{\textbf{V}}$ as the input to the VTF-E module. In the VTF-E module, we only have the word input for the VTF-E module without using the attention mechanism for visual features.
(2) \textbf{DaS(SA) w/o VF-D}: Our DaS approach without using visual features as the input in the decoder only.  As visual features are not used as the input for our VTF-D module, the attention mechanism for visual features in our VTF-D module is also not used, namely the VTF-D module only takes the textual information from each word and the attended hidden representation from the encoder-attention sub-net as the input.
(3) \textbf{DaS(SA) w/o VF-ED}: Our DaS approach without using visual features as the input of both encoder and decoder.

From the results in Table \ref{table:visual ablation}, we observe that our \textbf{DaS(SA)} outperforms \textbf{DaS(SA) w/o VF-E}, \textbf{Das(SA) w/o VF-D} and \textbf{DaS(SA) w/o VF-ED}, which demonstrates that it is beneficial to use the visual features as the input for both encoder and decoder and the attention mechanism for visual features.

 In Table \ref{table:lambda}, we evaluate the performance of our DaS when using different trade-off parameters $\lambda_d$ in Eq. (\ref{eq:loss}). We observe that our proposed method performs slightly better when setting $\lambda_{d}$ = 0.1. The Meteor score and Cider-D score decrease when we do not use the discriminative loss (i.e. $\lambda_{d} = 0$). These results demonstrate that it is useful to use our discriminative loss. In Table \ref{table:proposal_number}, we report the results of our method DaS when using different number of proposals, and our method achieves the best result when the number of proposals is set to 20. \color{black}

\begin{figure}[t]
    \centering
    \includegraphics[scale = .46]{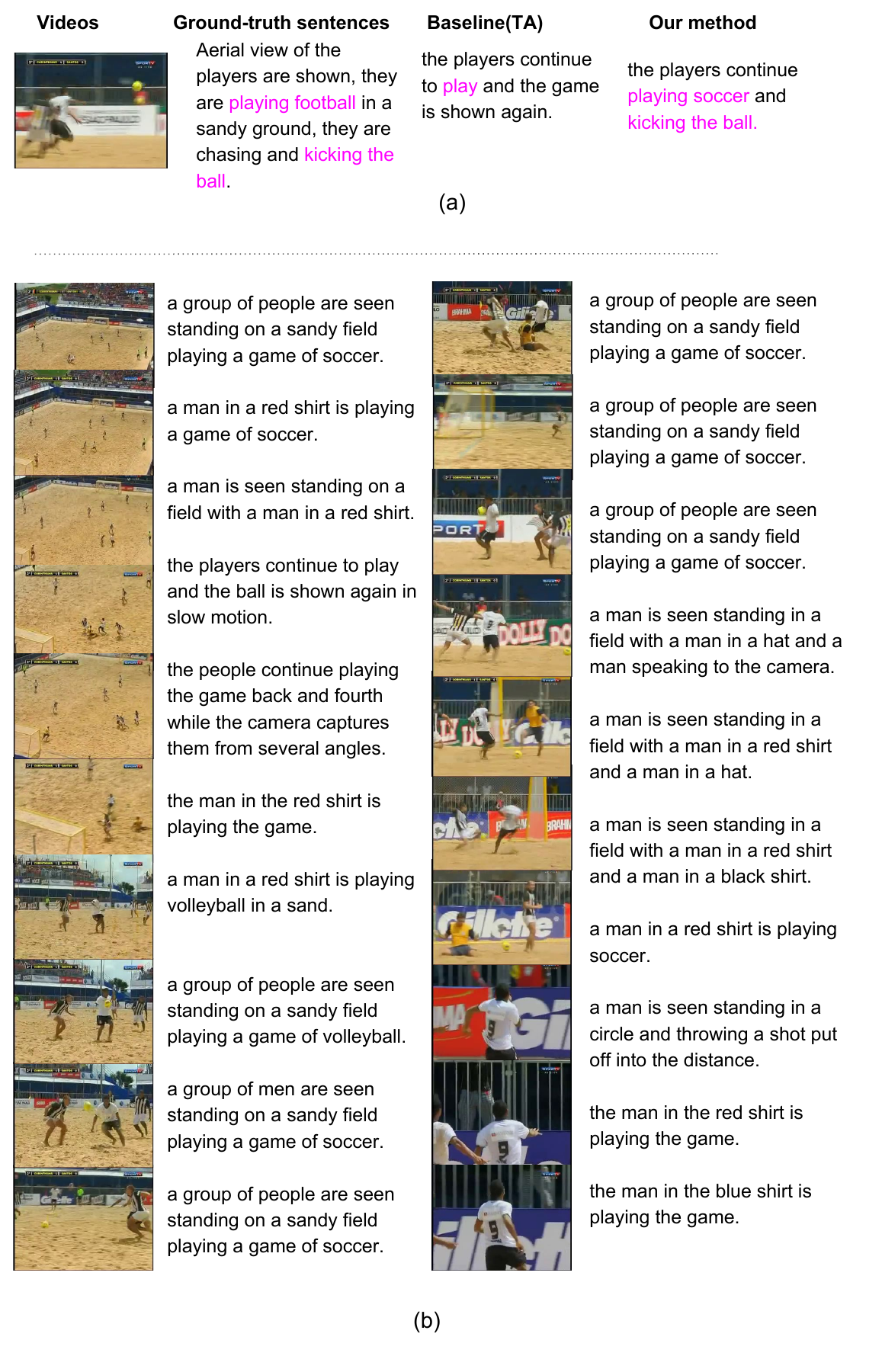}
    \vspace{-35pt}
    \caption{ (a) Comparison between our method \textbf{DaS(HA)} and the baseline method \textbf{TA} \cite{yao2015describing}. Correctly captured actions are highlighted by purple color (b) The output of our division module (i.e., $N_m$ = 20 sentences), which is also the textual input of our summarization module.}
    \label{visual3}
\end{figure}

\begin{figure}[t]
    \centering
    \includegraphics[scale = .64]{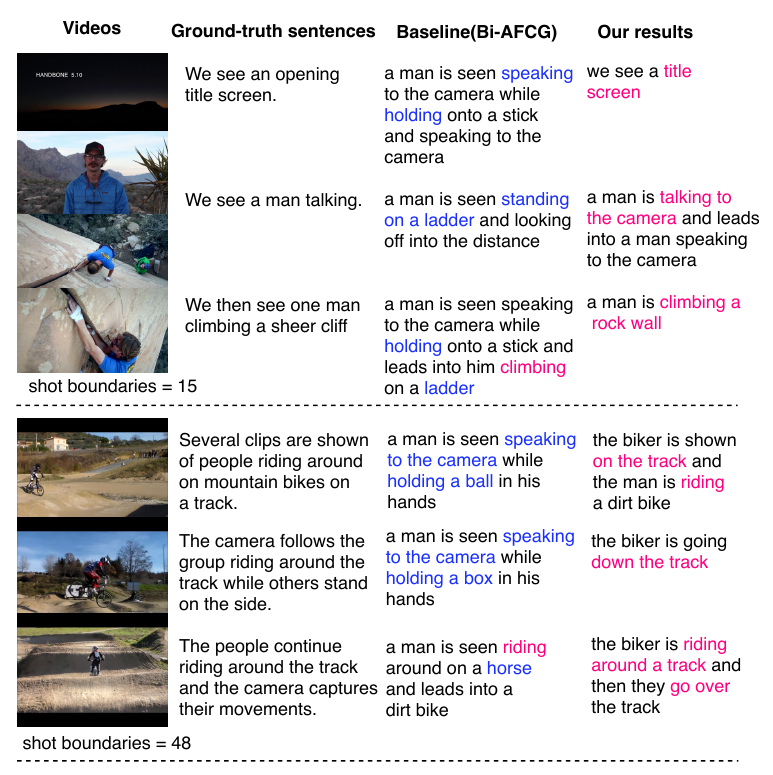}
    \vspace{-10pt}
    \caption{ Comparison between our method \textbf{DaS(HA)} and the baseline method \textbf{Bi-AFCG} \cite{wang2018bidirectional} for the videos that have large number of shot boundaries. Correctly captured actions are highlighted by purple color.}
    \label{shot_visual}
\end{figure}

\subsection{Qualitative Analysis}
In Figure \ref{qa}, we take one video to compare the generated sentences from our \textbf{DaS(HA)} and the baseline method \textbf{TA}\cite{yao2015describing}, which is implemented by ourself. We observe that we can generate better sentences for this video. 
For example, our method captures three key actions: ``raking leaves'', ``walking'' and ``speaking to the camera'', while the baseline method \textbf{TA}\cite{yao2015describing} only recognizes one action which is ``raking leaves'' in this video. 

In Figure \ref{visual2}, we show some generated sentences by using our method \textbf{DaS(HA)} based on the automatically generated proposals, in which we observe that our newly proposed method generates reasonable sentences for all videos.

In Figure \ref{visual3} (a), we show a representative video and compare the generated sentences by using our method \textbf{DaS(HA)} and the baseline method \textbf{TA} \cite{yao2015describing}. In Figure \ref{visual3} (b), we show $N_m=20$ sentences after using our division module for generating one sentence for each video segment, which is the output of our division module and also the textual input of our summarization module. While some sentences do not describe the correct action (e.g. the ground-truth action ``playing soccer''  is wrongly predicted as another action ``playing volleyball''), our work can generate much better sentence after summarizing all sentences by using our newly proposed summarization module.

 We show the qualitative analysis compared with \textbf{Bi-AFCG} method when the number of shot boundaries is large (the first video has 15 shot boundaries and the second video has 48 shot boundaries) in Fig. \ref{shot_visual}. Our method captures more correct actions while the baseline (\textbf{Bi-AFCG} in \cite{wang2018bidirectional}) only recognizes two actions ``standing'' and ``riding''. This result also shows that our method works well when the scenes and objects are rapidly changing.

\section{Conclusion}

By formulating the dense video captioning task as a new visual cue aided sentence summarization problem, we have proposed a new DaS framework in this work. In the division stage, we generate multiple sentence descriptions for describing diverse visual content for each event proposal. In the summarization stage, we propose a new two-stage LSTM network equipped with a new hierarchical attention mechanism, in which the first-stage LSTM network acts as the encoder to effectively fuse visual and textual features as a set of hidden representations, while the second-stage LSTM network equipped with a newly proposed hierarchical attention mechanism acts as the decoder to  generate one descriptive sentence. 
Extensive experiments on the ActivityNet Captions dataset demonstrate the effectiveness of our DaS framework for dense video captioning.

\ifCLASSOPTIONcaptionsoff
  \newpage
\fi



%


{\small
\bibliographystyle{IEEEtran}
\bibliography{main}

\begin{thebibliography}{10}
\providecommand{\url}[1]{#1}
\csname url@samestyle\endcsname
\providecommand{\newblock}{\relax}
\providecommand{\bibinfo}[2]{#2}
\providecommand{\BIBentrySTDinterwordspacing}{\spaceskip=0pt\relax}
\providecommand{\BIBentryALTinterwordstretchfactor}{4}
\providecommand{\BIBentryALTinterwordspacing}{\spaceskip=\fontdimen2\font plus
\BIBentryALTinterwordstretchfactor\fontdimen3\font minus \fontdimen4\font\relax}
\providecommand{\BIBforeignlanguage}[2]{{%
\expandafter\ifx\csname l@#1\endcsname\relax
\typeout{** WARNING: IEEEtran.bst: No hyphenation pattern has been}%
\typeout{** loaded for the language `#1'. Using the pattern for}%
\typeout{** the default language instead.}%
\else
\language=\csname l@#1\endcsname
\fi
#2}}
\providecommand{\BIBdecl}{\relax}
\BIBdecl

\bibitem{krishna2017dense}
R.~Krishna, K.~Hata, F.~Ren, L.~Fei-Fei, and J.~C. Niebles, ``Dense-captioning events in videos.'' in \emph{ICCV}, 2017, pp. 706--715.

\bibitem{wang2018bidirectional}
J.~Wang, W.~Jiang, L.~Ma, W.~Liu, and Y.~Xu, ``Bidirectional attentive fusion with context gating for dense video captioning,'' in \emph{Proceedings of the IEEE Conference on Computer Vision and Pattern Recognition}, 2018, pp. 7190--7198.

\bibitem{li2018jointly}
Y.~Li, T.~Yao, Y.~Pan, H.~Chao, and T.~Mei, ``Jointly localizing and describing events for dense video captioning,'' in \emph{Proceedings of the IEEE Conference on Computer Vision and Pattern Recognition}, 2018, pp. 7492--7500.

\bibitem{zhou2018end}
L.~Zhou, Y.~Zhou, J.~J. Corso, R.~Socher, and C.~Xiong, ``End-to-end dense video captioning with masked transformer,'' in \emph{Proceedings of the IEEE Conference on Computer Vision and Pattern Recognition}, 2018, pp. 8739--8748.

\bibitem{xu2015show}
K.~Xu, J.~Ba, R.~Kiros, K.~Cho, A.~Courville, R.~Salakhudinov, R.~Zemel, and Y.~Bengio, ``Show, attend and tell: Neural image caption generation with visual attention,'' in \emph{International conference on machine learning}, 2015, pp. 2048--2057.

\bibitem{duchenne2009automatic}
O.~Duchenne, I.~Laptev, J.~Sivic, F.~Bach, and J.~Ponce, ``Automatic annotation of human actions in video,'' in \emph{Computer Vision, 2009 IEEE 12th International Conference on}.\hskip 1em plus 0.5em minus 0.4em\relax IEEE, 2009, pp. 1491--1498.

\bibitem{caba2016fast}
F.~Caba~Heilbron, J.~Carlos~Niebles, and B.~Ghanem, ``Fast temporal activity proposals for efficient detection of human actions in untrimmed videos,'' in \emph{Proceedings of the IEEE conference on computer vision and pattern recognition}, 2016, pp. 1914--1923.

\bibitem{escorcia2016daps}
V.~Escorcia, F.~C. Heilbron, J.~C. Niebles, and B.~Ghanem, ``Daps: Deep action proposals for action understanding,'' in \emph{European Conference on Computer Vision}.\hskip 1em plus 0.5em minus 0.4em\relax Springer, 2016, pp. 768--784.

\bibitem{buch2017sst}
S.~Buch, V.~Escorcia, C.~Shen, B.~Ghanem, and J.~C. Niebles, ``Sst: Single-stream temporal action proposals,'' in \emph{2017 IEEE Conference on Computer Vision and Pattern Recognition (CVPR)}.\hskip 1em plus 0.5em minus 0.4em\relax IEEE, 2017, pp. 6373--6382.

\bibitem{shen2017weakly}
Z.~Shen, J.~Li, Z.~Su, M.~Li, Y.~Chen, Y.-G. Jiang, and X.~Xue, ``Weakly supervised dense video captioning,'' in \emph{The IEEE Conference on Computer Vision and Pattern Recognition (CVPR)}, vol.~2, no.~7, 2017.

\bibitem{pan2016jointly}
Y.~Pan, T.~Mei, T.~Yao, H.~Li, and Y.~Rui, ``Jointly modeling embedding and translation to bridge video and language,'' in \emph{Proceedings of the IEEE conference on computer vision and pattern recognition}, 2016, pp. 4594--4602.

\bibitem{chen2018less}
Y.~Chen, S.~Wang, W.~Zhang, and Q.~Huang, ``Less is more: Picking informative frames for video captioning,'' \emph{arXiv preprint arXiv:1803.01457}, 2018.

\bibitem{wang2018video}
X.~Wang, W.~Chen, J.~Wu, Y.-F. Wang, and W.~Y. Wang, ``Video captioning via hierarchical reinforcement learning,'' in \emph{Proceedings of the IEEE Conference on Computer Vision and Pattern Recognition}, 2018, pp. 4213--4222.

\bibitem{wang2018reconstruction}
B.~Wang, L.~Ma, W.~Zhang, and W.~Liu, ``Reconstruction network for video captioning,'' in \emph{Proceedings of the IEEE Conference on Computer Vision and Pattern Recognition}, 2018, pp. 7622--7631.

\bibitem{wu2018interpretable}
X.~Wu, G.~Li, Q.~Cao, Q.~Ji, and L.~Lin, ``Interpretable video captioning via trajectory structured localization,'' in \emph{Proceedings of the IEEE Conference on Computer Vision and Pattern Recognition}, 2018, pp. 6829--6837.

\bibitem{chen2018regularizing}
X.~Chen, L.~Ma, W.~Jiang, J.~Yao, and W.~Liu, ``Regularizing rnns for caption generation by reconstructing the past with the present,'' \emph{arXiv preprint arXiv:1803.11439}, 2018.

\bibitem{su2019improving}
R.~Su, W.~Ouyang, L.~Zhou, and D.~Xu, ``Improving action localization by progressive cross-stream cooperation,'' in \emph{Proceedings of the IEEE Conference on Computer Vision and Pattern Recognition}, 2019, pp. 12\,016--12\,025.

\bibitem{venugopalan2014translating}
S.~Venugopalan, H.~Xu, J.~Donahue, M.~Rohrbach, R.~Mooney, and K.~Saenko, ``Translating videos to natural language using deep recurrent neural networks,'' \emph{arXiv preprint arXiv:1412.4729}, 2014.

\bibitem{gan2017semantic}
Z.~Gan, C.~Gan, X.~He, Y.~Pu, K.~Tran, J.~Gao, L.~Carin, and L.~Deng, ``Semantic compositional networks for visual captioning,'' in \emph{Proceedings of the IEEE Conference on Computer Vision and Pattern Recognition}, vol.~2, 2017.

\bibitem{yao2015describing}
L.~Yao, A.~Torabi, K.~Cho, N.~Ballas, C.~Pal, H.~Larochelle, and A.~Courville, ``Describing videos by exploiting temporal structure,'' in \emph{Proceedings of the IEEE international conference on computer vision}, 2015, pp. 4507--4515.

\bibitem{pan2017video}
Y.~Pan, T.~Yao, H.~Li, and T.~Mei, ``Video captioning with transferred semantic attributes,'' in \emph{CVPR}, vol.~2, 2017, p.~3.

\bibitem{sutskever2014sequence}
I.~Sutskever, O.~Vinyals, and Q.~V. Le, ``Sequence to sequence learning with neural networks,'' in \emph{Advances in neural information processing systems}, 2014, pp. 3104--3112.

\bibitem{donahue2015long}
J.~Donahue, L.~Anne~Hendricks, S.~Guadarrama, M.~Rohrbach, S.~Venugopalan, K.~Saenko, and T.~Darrell, ``Long-term recurrent convolutional networks for visual recognition and description,'' in \emph{Proceedings of the IEEE conference on computer vision and pattern recognition}, 2015, pp. 2625--2634.

\bibitem{pan2016hierarchical}
P.~Pan, Z.~Xu, Y.~Yang, F.~Wu, and Y.~Zhuang, ``Hierarchical recurrent neural encoder for video representation with application to captioning,'' in \emph{Proceedings of the IEEE Conference on Computer Vision and Pattern Recognition}, 2016, pp. 1029--1038.

\bibitem{li2015summarization}
G.~Li, S.~Ma, and Y.~Han, ``Summarization-based video caption via deep neural networks,'' in \emph{Proceedings of the 23rd ACM international conference on Multimedia}.\hskip 1em plus 0.5em minus 0.4em\relax ACM, 2015, pp. 1191--1194.

\bibitem{liu2016boosting}
Y.~Liu and Z.~Shi, ``Boosting video description generation by explicitly translating from frame-level captions,'' in \emph{Proceedings of the 24th ACM international conference on Multimedia}.\hskip 1em plus 0.5em minus 0.4em\relax ACM, 2016, pp. 631--634.

\bibitem{erkan2004lexrank}
G.~Erkan and D.~R. Radev, ``Lexrank: Graph-based lexical centrality as salience in text summarization,'' \emph{Journal of artificial intelligence research}, vol.~22, pp. 457--479, 2004.

\bibitem{xu2016ask}
H.~Xu and K.~Saenko, ``Ask, attend and answer: Exploring question-guided spatial attention for visual question answering,'' in \emph{European Conference on Computer Vision}.\hskip 1em plus 0.5em minus 0.4em\relax Springer, 2016, pp. 451--466.

\bibitem{song2017end}
S.~Song, C.~Lan, J.~Xing, W.~Zeng, and J.~Liu, ``An end-to-end spatio-temporal attention model for human action recognition from skeleton data.'' in \emph{AAAI}, vol.~1, no.~2, 2017, pp. 4263--4270.

\bibitem{you2016image}
Q.~You, H.~Jin, Z.~Wang, C.~Fang, and J.~Luo, ``Image captioning with semantic attention,'' in \emph{Proceedings of the IEEE conference on computer vision and pattern recognition}, 2016, pp. 4651--4659.

\bibitem{yang2016stacked}
Z.~Yang, X.~He, J.~Gao, L.~Deng, and A.~Smola, ``Stacked attention networks for image question answering,'' in \emph{Proceedings of the IEEE Conference on Computer Vision and Pattern Recognition}, 2016, pp. 21--29.

\bibitem{wang2017residual}
F.~Wang, M.~Jiang, C.~Qian, S.~Yang, C.~Li, H.~Zhang, X.~Wang, and X.~Tang, ``Residual attention network for image classification,'' \emph{arXiv preprint arXiv:1704.06904}, 2017.

\bibitem{ba2014multiple}
J.~Ba, V.~Mnih, and K.~Kavukcuoglu, ``Multiple object recognition with visual attention,'' \emph{arXiv preprint arXiv:1412.7755}, 2014.

\bibitem{vinyals2015show}
O.~Vinyals, A.~Toshev, S.~Bengio, and D.~Erhan, ``Show and tell: A neural image caption generator,'' in \emph{Proceedings of the IEEE conference on computer vision and pattern recognition}, 2015, pp. 3156--3164.

\bibitem{hochreiter1997long}
S.~Hochreiter and J.~Schmidhuber, ``Long short-term memory,'' \emph{Neural computation}, vol.~9, no.~8, pp. 1735--1780, 1997.

\bibitem{yang2016review}
Z.~Yang, Y.~Yuan, Y.~Wu, W.~W. Cohen, and R.~R. Salakhutdinov, ``Review networks for caption generation,'' in \emph{Advances in Neural Information Processing Systems}, 2016, pp. 2361--2369.

\bibitem{rennie2017self}
S.~J. Rennie, E.~Marcheret, Y.~Mroueh, J.~Ross, and V.~Goel, ``Self-critical sequence training for image captioning,'' in \emph{CVPR}, vol.~1, no.~2, 2017, p.~3.

\bibitem{venugopalan2015sequence}
S.~Venugopalan, M.~Rohrbach, J.~Donahue, R.~Mooney, T.~Darrell, and K.~Saenko, ``Sequence to sequence-video to text,'' in \emph{Proceedings of the IEEE international conference on computer vision}, 2015, pp. 4534--4542.

\bibitem{yu2016video}
H.~Yu, J.~Wang, Z.~Huang, Y.~Yang, and W.~Xu, ``Video paragraph captioning using hierarchical recurrent neural networks,'' in \emph{Proceedings of the IEEE conference on computer vision and pattern recognition}, 2016, pp. 4584--4593.

\bibitem{caba2015activitynet}
F.~Caba~Heilbron, V.~Escorcia, B.~Ghanem, and J.~Carlos~Niebles, ``Activitynet: A large-scale video benchmark for human activity understanding,'' in \emph{Proceedings of the IEEE Conference on Computer Vision and Pattern Recognition}, 2015, pp. 961--970.

\bibitem{kingma2014adam}
D.~P. Kingma and J.~Ba, ``Adam: A method for stochastic optimization,'' \emph{arXiv preprint arXiv:1412.6980}, 2014.

\bibitem{papineni2002bleu}
K.~Papineni, S.~Roukos, T.~Ward, and W.-J. Zhu, ``Bleu: a method for automatic evaluation of machine translation,'' in \emph{Proceedings of the 40th annual meeting on association for computational linguistics}.\hskip 1em plus 0.5em minus 0.4em\relax Association for Computational Linguistics, 2002, pp. 311--318.

\bibitem{lin2004rouge}
C.-Y. Lin, ``Rouge: A package for automatic evaluation of summaries,'' in \emph{Text summarization branches out: Proceedings of the ACL-04 workshop}, vol.~8.\hskip 1em plus 0.5em minus 0.4em\relax Barcelona, Spain, 2004.

\bibitem{denkowski2014meteor}
M.~Denkowski and A.~Lavie, ``Meteor universal: Language specific translation evaluation for any target language,'' in \emph{Proceedings of the ninth workshop on statistical machine translation}, 2014, pp. 376--380.

\bibitem{vedantam2015cider}
R.~Vedantam, C.~Lawrence~Zitnick, and D.~Parikh, ``Cider: Consensus-based image description evaluation,'' in \emph{Proceedings of the IEEE conference on computer vision and pattern recognition}, 2015, pp. 4566--4575.

\end{thebibliography}
}

%

\vspace{-15mm}
\begin{IEEEbiography}[{\includegraphics[width=1in,height=1.25in,clip,keepaspectratio]{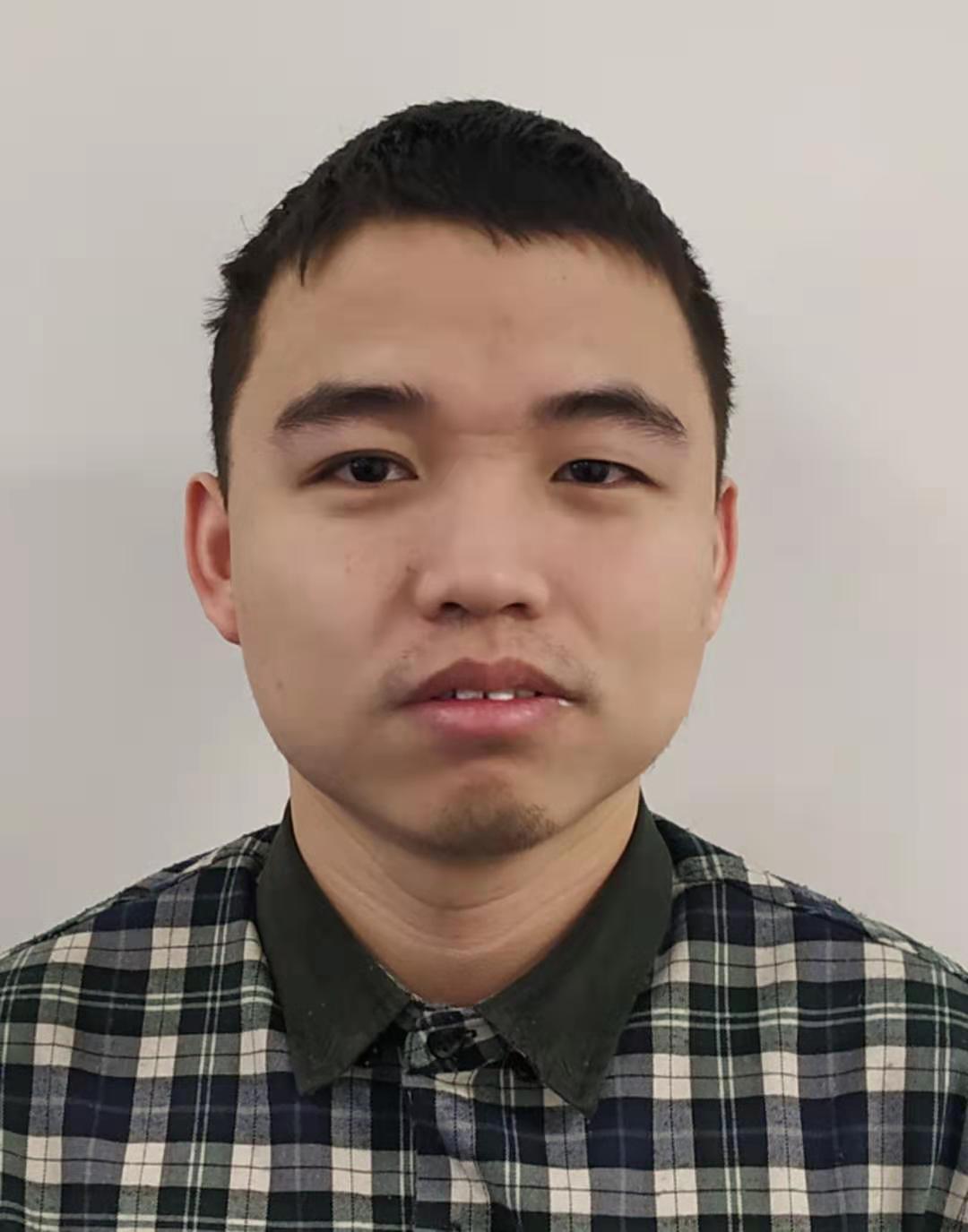}}]{Zhiwang Zhang}
received BE degree in school of Electrical and Information Engineering from the University of Sydney in 2017, where he is currently pursuing the PhD degree. His research interests include deep learning and its applications on computer vision and natural language processing. 
\end{IEEEbiography}

\vspace{-15mm}
\begin{IEEEbiography}[{\includegraphics[width=1in,height=1.25in,clip,keepaspectratio]{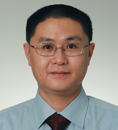}}]{Dong Xu}
 received the BE and PhD degrees from University of Science and Technology of China, in 2001 and 2005, respectively. While pursuing the PhD degree, he was an intern with Microsoft Research Asia, Beijing, China, and a research assistant with the Chinese University of Hong Kong, Shatin, Hong Kong, for more than two years. He was a post-doctoral research scientist with Columbia University, New York, NY, for one year. He worked as a faculty member with Nanyang Technological University, Singapore. Currently, he is a professor and chair in Computer Engineering with the School of Electrical and Information Engineering, the University of Sydney, Australia. His current research interests include computer vision, statistical learning, and multimedia content analysis. He was the co-author of a paper that won the Best Student Paper award in the IEEE Conferenceon Computer Vision and Pattern Recognition (CVPR) in 2010, and a paper that won the Prize Paper award in IEEE Transactions on Multimedia (T-MM) in 2014. He is a fellow of the IEEE.
\end{IEEEbiography}
\vspace{-15mm}
\begin{IEEEbiography}[{\includegraphics[width=1in,height=1.25in,clip,keepaspectratio]{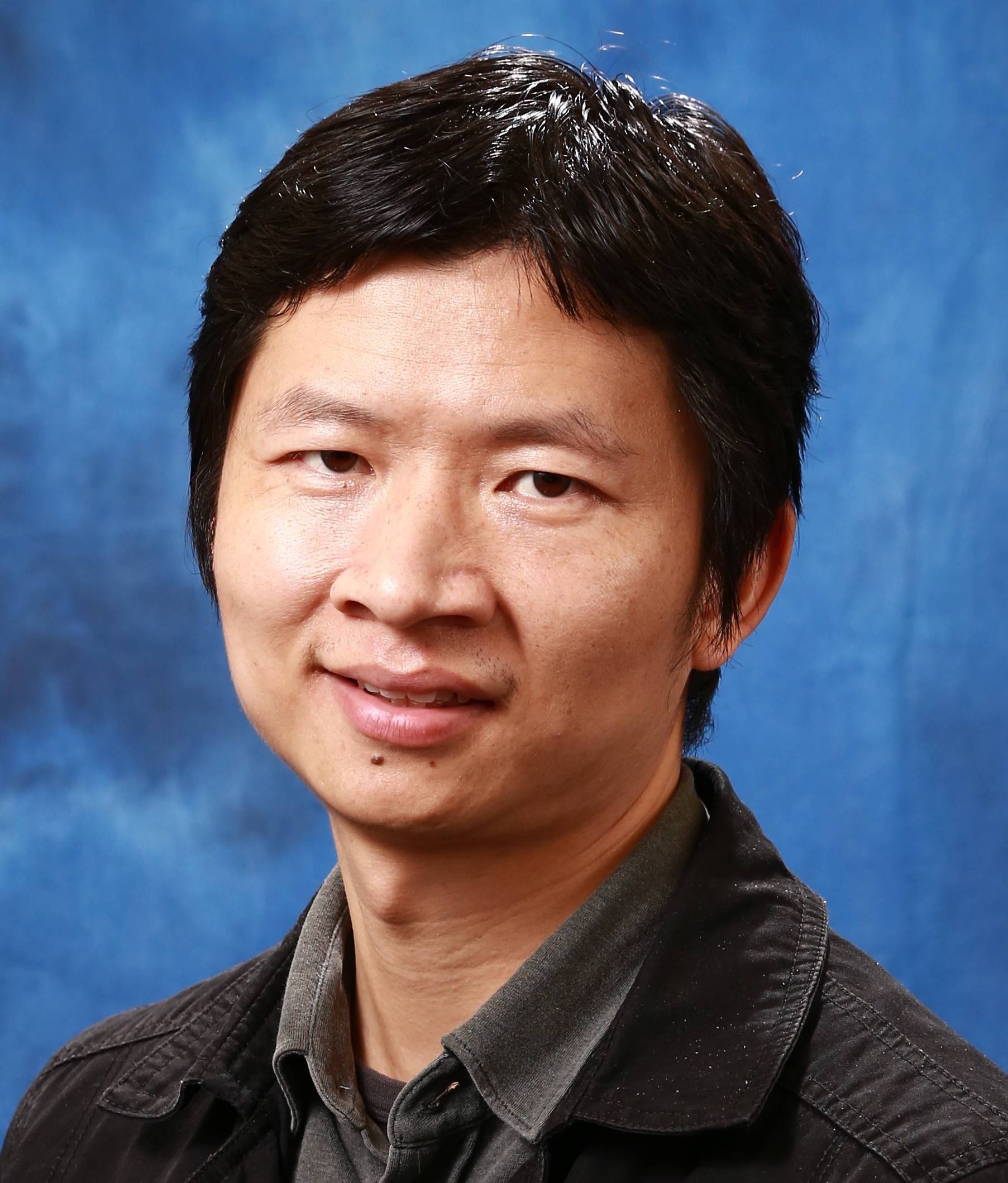}}]{Wanli Ouyang}
received the PhD degree in the Department of Electronic Engineering, Chinese University of Hong Kong. Since 2017, he is a senior lecturer with the University of Sydney. His research interests include image processing, computer vision, and pattern recognition. He is a senior member of the IEEE.
\end{IEEEbiography}
\vspace{-15mm}
\begin{IEEEbiography}[{\includegraphics[width=1in,height=1.25in,clip,keepaspectratio]{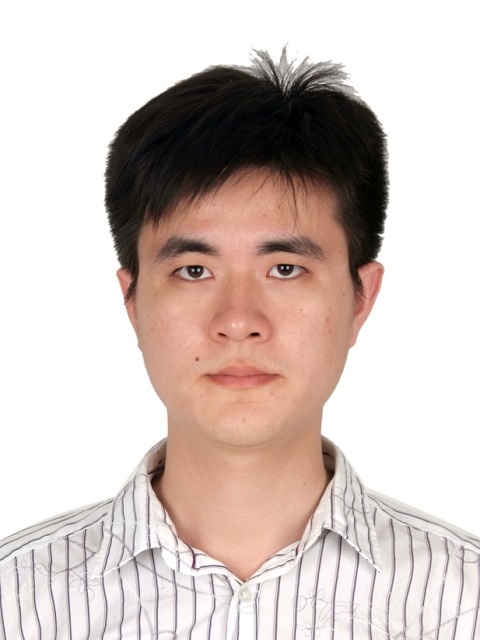}}]{Chuanqi Tan}
received Ph.D. degree in computer sciences and technology from Tsinghua University in July 2019. Before this, he was worked at baidu.com in 2014 and jike.com in 2012. His research interests include artificial intelligence, deep learning, transfer learning and brain-computer.
\end{IEEEbiography}







\end{document}